\documentclass[letterpaper]{article} 
\usepackage{aaai24}  
\usepackage{times}  
\usepackage{helvet}  
\usepackage{courier}  
\usepackage[hyphens]{url}  
\usepackage{graphicx} 
\usepackage{natbib}  
\usepackage{caption} 
\usepackage{algorithm}
\usepackage{cite}
\usepackage{amsmath,amssymb,amsfonts}
\usepackage{textcomp}
\usepackage[table]{xcolor}
\usepackage{bm}
\usepackage{booktabs}
\usepackage{multirow}
\usepackage{makecell}
\usepackage{subfig}
\usepackage{afterpage}
\usepackage{algpseudocode}
\usepackage{rotating}
\usepackage{nicematrix}
\usepackage{capt-of}
\usepackage{xfrac}
\usepackage{mathtools}
\usepackage{bookmark}

\usepackage{newfloat}
\usepackage{listings}
\DeclareCaptionStyle{ruled}{labelfont=normalfont,labelsep=colon,strut=off} 
\floatstyle{ruled}
\newfloat{listing}{tb}{lst}{}
\floatname{listing}{Listing}
\DeclareMathOperator{\mc}{mc} 
\DeclareMathOperator{\IV}{IV} 
\DeclareMathOperator{\sigmoid}{sigmoid} 
\DeclarePairedDelimiter{\nfloor}{\lfloor}{\rfloor} 
\makeatletter
\DeclareRobustCommand{\volume}{\text{\volumedash}V}
\newcommand{\volumedash}{%
  \makebox[0pt][l]{%
    \ooalign{\hfil\hphantom{$\m@th V$}\hfil\cr\kern0.08em--\hfil\cr}%
  }%
}
\makeatother

\usepackage{xspace}
\makeatletter
\DeclareRobustCommand\onedot{\futurelet\@let@token\@onedot}
\def\@onedot{\ifx\@let@token.\else.\null\fi\xspace}

\def\eg{\emph{e.g}\onedot} \def\Eg{\emph{E.g}\onedot}
\def\ie{\emph{i.e}\onedot} \def\Ie{\emph{I.e}\onedot}
\def\cf{\emph{c.f}\onedot} 
\def\etc{\emph{etc}\onedot} 
\def\wrt{w.r.t\onedot}

\newcommand{\beginsupplement}{%
        \setcounter{table}{0}
        \renewcommand{\thetable}{S\arabic{table}}%
        \setcounter{figure}{0}
        \renewcommand{\thefigure}{S\arabic{figure}}%
     }

\title{Implicit Modeling of Non-Rigid Objects with Cross-Category Signals}
\author {
Yuchun Liu, 
Benjamin Planche, 
Meng Zheng, 
Zhongpai Gao, 
Pierre Sibut-Bourde, 
Fan Yang, 
Terrence Chen, 
Ziyan Wu 
}
\affiliations {
    United Imaging Intelligence\\
    {\scriptsize\texttt{\{yuchun.lui01, benjamin.planche, meng.zheng, zhongpai.gao, fan.yang03, terrence.chen, ziyan.wu\}@uii-ai.com}}
}
\date{May 2023}

\begin{document}

\twocolumn[{%
\renewcommand\twocolumn[1][]{#1}%
\maketitle
\begin{center}
    \centering
    \captionsetup{type=figure}
    \includegraphics[width=1.\textwidth]{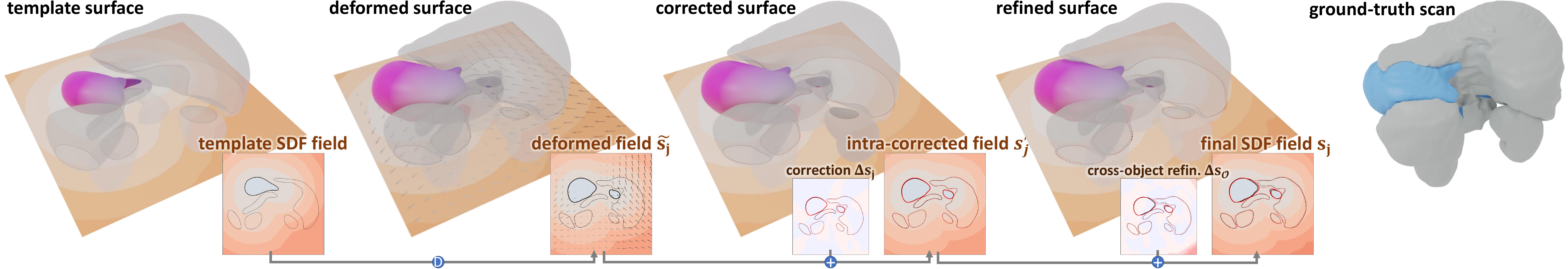}
    \captionof{figure}{Visualization of intermediary SDF slices (corresponding to the colored \texttt{stomach} class) and resulting 3D shapes predicted by our method. 
    Tackling the modeling of multi-object 3D instances, our neural SDF model goes beyond prior work by accounting for object interrelations (both for inference and supervision) to refine its predictions (better viewed zoomed in).}
    \label{fig:ablation_study}
    \vspace{1em}
\end{center}%
}]

\insert\footins{\noindent\footnotesize
Copyright \copyright \space 2024,
Association for the Advancement of Artificial Intelligence (www.aaai.org).
All rights reserved.}

\begin{abstract}
 Deep implicit functions (DIFs) have emerged as a potent and articulate means of representing 3D shapes. However, methods modeling object categories or non-rigid entities have mainly focused on single-object scenarios. In this work, we propose MODIF, a multi-object deep implicit function that jointly learns the deformation fields and instance-specific latent codes for multiple objects at once. Our emphasis is on non-rigid, non-interpenetrating entities such as organs. To effectively capture the interrelation between these entities and ensure precise, collision-free representations, our approach facilitates signaling between category-specific fields to adequately rectify shapes. We also introduce novel inter-object supervision: an attraction-repulsion loss is formulated to refine contact regions between objects. Our approach is demonstrated on various medical benchmarks, involving modeling different groups of intricate anatomical entities. Experimental results illustrate that our model can proficiently learn the shape representation of each organ and their relations to others, to the point that shapes missing from unseen instances can be consistently recovered by our method. Finally, MODIF can also propagate semantic information throughout the population via accurate point correspondences.
\end{abstract}

\section{Introduction}
In recent years, there has been extensive research into deep implicit functions (DIFs) as a means of representing 3D object categories. In comparison to conventional geometric representations, DIFs exhibit potential in effectively reconstructing geometric intricacies, even for non-rigid objects or object categories with large variability. This potential holds significant advantages for various clinical imaging applications, such as 3D organ reconstruction, anatomical segmentation, and surgical navigation. 
For example, various medical procedures depend on accurately locating precise organ regions, which implies accurately modeling the whole anatomical structure.

However, methods proposed so far to represent 3D entities \cite{park2019deepsdf,mescheder2019occupancy,sitzmann2020implicit,sun2022topology} are constrained to single-object scenarios.  
The exploration of multi-object 3D neural representations remains relatively uncharted, and prior research \cite{zhang2022implicit} tends to overlook the interrelations between the different categories that constitute each instance.
\Eg, for organs, their shape and location depends on a variety of criteria such as body pose\cite{guo2022smpl}, metabolic cycles, \etc. 
More importantly, their shapes undergo deformation based on the interactions and pressure they exert on one another. In this context, learning the implicit function of each category in isolation is both ineffective and problematic, as it could potentially lead to shape interpenetration. 

Dealing with multiple objects presents several challenges. First, the rigidity and shape variability across different object categories adds complexity to the task of simultaneously modeling various types of surface deformation fields. This challenge is often exacerbated by the scarcity of training data, particularly in the context of medical applications. This scarcity makes the task of achieving shape generalization even more complex.
Secondly, the model must jointly account for the interactions among objects. This involves not only capturing their relative positions but also modeling the contact regions occurring between these objects.



To address these challenges, we introduce our Multi-Object Deformed Implicit Field (MODIF) model, designed to learn implicit multi-object shape functions. It incorporates a cross-category refinement mechanism that allows us to capture interactions between objects while still maintaining the accuracy of individual reconstructions. The model facilitates the generation of both point correspondences and separate per-category templates, which can be utilized for extrapolating the shape of a specific object missing in an unseen instance.

In summary, the primary contributions of our paper can be outlined as follows:

\begin{itemize}
\item We introduce a comprehensive method for representing multiple non-rigid objects using an implicit approach. Our model not only produces accurate reconstructions of shapes but also offers precise predictions of individual object positions. 

\item We design a cross-category refinement mechanism, incorporating the features from each individual sub-function to generate an overall correction field. With this, an attraction-repulsion loss is formulated to supervise contact regions between objects and to effectively reduce erroneous object interpenetration.  

\item Our model can generate point correspondences for multiple objects simultaneously. We ensure multi-category point correspondence while preserving single-category geometries. 

\item We provide a solution for reconstructing plausible and consistent shapes when a specific object is missing for new observed instances. 

\item We evaluate our solution on 3 different datasets over multiple tasks and show that MODIF consistently outperforms state-of-the-art methods, even when the latter are provided with additional supervision.

\end{itemize}

\section{Related works}
\paragraph{Deep Implicit Functions.}
DIF methods rely on neural networks to represent continuous 3D shapes and are nowadays considered more flexible and efficient than traditional explicit methods. 
There has been a proliferation of models \cite{hao2020dualsdf,duan2020curriculum,chabra2020deep} developed in this direction. 
DeepSDF \cite{park2019deepsdf} introduces an auto-decoder model to learn a 3D signed distance field (SDF). Occupancy networks \cite{mescheder2019occupancy} are another popular method, introducing a deep neural network classifier to decide if the point is inside of the object. It inspired various following works \cite{peng2020convolutional}\cite{roddick2020predicting}\cite{lionar2021dynamic}.

Despite the impressive shape representation capabilities of DIFs, the reconstruction outcomes often suffer from a lack of intricate details. In response to this issue, \cite{sitzmann2020implicit} introduced a periodic activation function to model finer details and improve the overall accuracy over complex scenes.

\paragraph{Template Learning and Point Correspondence.}
In medical imaging applications, the establishment of point correspondences holds significant importance, as it often becomes necessary to map and compare various anatomical structures. To address this need, several deformation-based DIF methods have been introduced to infer dense correspondences across shapes. 
Deep Implicit Templates (DIT) \cite{zheng2021deep} utilize an LSTM model to learn conditional deformations and generate templates spanning all shapes. 
Built upon, Neural Diffeomorphic Flow (NDF) \cite{sun2022topology} employs multiple neural ordinary equation (NODE) blocks to ensure the preservation of topological features during shape deformation; whereas Deformed Implicit Field (DIF) \cite{deng2021deformed} offers greater generalization capability. It is worth noting that these methods are tailored for single-object scenarios, while our approach is capable of concurrently generating point correspondences for multiple objects.

\paragraph{Structured Shape Representation.} 
To capture complex 3D shapes, recent endeavors have focused on breaking down instances into simpler, smaller components. Notably, DeepLS \cite{chabra2020deep} uses a grid of independent latent codes to model local structures; and 
LDIF \cite{genova2020local} partitions the 3D space into a structured arrangement of learned implicit functions. 
DMM model \cite{zhang2022implicit} presents an implicit dental model, which provides the segmentation labels for individual teeth and gum.
Although ImgHUM and DMM are capable of generating distinct sub-parts, their approaches primarily focus on reconstructing the entire instance rather than comprehensively modeling relationships between these sub-parts. In contrast, our model is specifically designed to address interactions and is thus proficient in circumventing collision problems between objects.


\begin{figure*}[t]
    \centering
    \includegraphics[width=1.\textwidth]{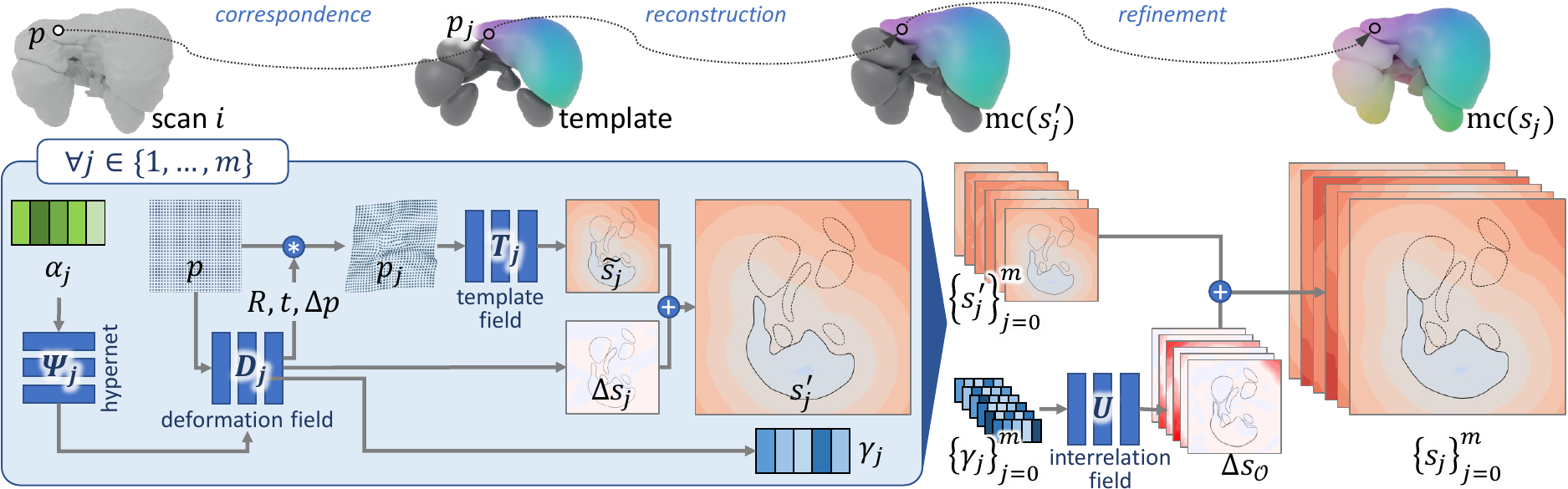}
    \caption{Proposed pipeline, composed of $m$ sub-functions separately modeling the categories composing target 3D instances, and one interrelational refinement field. It achieves accurate instance reconstruction, point correspondence, and object recovery.}
    \label{fig:pipeline}
\end{figure*}

\section{Method}
\subsection{Overview}

Our goal is to learn a modular shape representation for a collection $\{\mathcal{O}_i\}_{i=1}^n$ of 3D instances, with each instance $\mathcal{O}_i$ being composed of a set of interrelated and non-overlapping objects $\{\mathcal{O}_{i,j}\}_{j=1}^m$ belonging to $m$ different categories. 
We define a function $F(\alpha_i) = \mathcal{O}_i$ that maps a latent vector $\alpha_i$ to the corresponding 3D instance, and we jointly learn $F$ and $\{\alpha_i\} \in \mathbb{R}^k$ in a self-supervised manner.
Similar to previous works \cite{park2019deepsdf,alldieck2021imghum,sun2022topology,deng2021deformed,zhang2022implicit}, we use signed distance fields (SDFs), which can continuously and implicitly represent surface geometries and be modeled by coordinate-based neural networks. For every point $p \in \mathbb{R}^3$ in the represented domain, an SDF provides a scalar value $s$ corresponding to the distance from $p$ to the closest object surface, with $s<0$ if $p$ is inside the object and $s>0$ if $p$ is outside. 
The explicit surface representation of each shape can be defined as the zero-level set of its SDF, \eg, which can be extracted using the marching-cubes algorithm (noted $\mc$) \cite{lorensen1998marching}.
Therefore, our function $F$ can be expressed as:
\begin{equation}
    F = \mc\left(\left\{f(\alpha_i, p) \mid p \sim \Omega \right\}\right),
\end{equation}
with $f: \mathbb{R}^k \times \mathbb{R}^3 \mapsto \mathbb{R}^m$ a novel neural implicit model that, given a point $p$ in the 3D space $\Omega$ and a conditioning vector $\alpha_i$ learned from $\mathcal{O}_i$, predicts $s_i = \{s_{i,j}\}_{j=1}^m$, a vector of SDF values \wrt each of the $m$ instance shapes.

This formulation differs from most prior arts that output a single distance field per instance. When considering instances with multiple objects, they either (a) merge their geometries into one, prior to modeling it as a single field \cite{sun2022topology,zhang2022implicit}; or (b) tackle each object separately, each with their own distinct and uncorrelated latent shape space \cite{deng2021deformed}.
In this work, we aim at learning object-compositional implicit functions that account for interrelations between the $m$ object categories to ensure higher inter- and cross-instance consistency.  
Inspired by \cite{alldieck2021imghum,zhang2022implicit}, we decompose the problem into learning $m$ sub-functions $f_j$: 
\begin{equation}
    f_j: (\alpha_{i,j}, p) \in \mathbb{R}^{\nfloor{\sfrac{k}{m}}} \times \mathbb{R}^3 \mapsto (s'_{i,j}, \gamma_{i,j}) \in \mathbb{R} \times \mathbb{R}^l,
\end{equation}
with $s'_{i,j} \in \mathbb{R}$ the SDF value of $p$ \wrt $\mathcal{O}_{i,j}$ estimated by this sub-function and $\gamma_{i,j}$ an $l$-dimensional feature vector encoding geometrical properties of shape $\mathcal{O}_{i,j}$.
Unlike previous work, we encourage signaling between these sub-functions, both in a feed-forward (feature sharing \cf proposed $\gamma_{i,j}$) and back-propagation (cross-category supervision) manners. 
We further introduce a cross-category refinement neural model $U$, defined as follows: 
\begin{equation}
U: \left(\{\gamma_{i,j}\}_{j=1}^m\right) \in \mathbb{R}^{l \times m} \mapsto \Delta s_{\mathcal{O}_i} \in \mathbb{R}^m,    
\end{equation}
\Ie, it generates a $m$-dimensional residual SDF vector to consolidate and correct the per-module SDF predictions using cross-object signals $\{\gamma_{i,j}\}_{j=1}^m$. 
The refined SDF predictions thus correspond to $s_i = s'_i + \Delta s_{\mathcal{O}_i}$.
In the remainder of this section, we further detail our model, its supervision, and application to multi-object reconstruction and point correspondence (as shown in Figure \ref{fig:pipeline}).

\subsection{Network Structure}


Our sub-functions $f_j$ are adapted from DIF-Net \cite{deng2021deformed} and composed of 3 neural networks---neural template field $T_j$, hyper-net $\Psi_j$, and dual deformation/correction field $D_j$---to predict $s_{i,j}$ based on instance-specific learned sub-code $\alpha_{i,j}$. The predicted $m$-dimensional SDF vector is then residually edited by our novel cross-object correction function $U$.


\paragraph{Template Field $T_j$.}
Defined as $T_j: p \in \mathbb{R}^3 \mapsto \widetilde{s_j} \in \mathbb{R}$, it maps 3D points to their signed distances \wrt template object $j$ in a reference space. As proposed in \cite{deng2021deformed}, its neural weights are shared across the whole population, \ie, to learn category-wide shape properties.

\paragraph{Generalized Deformation/Correction Field $D_{\Psi_j
}$.}
DIF-Net authors propose a function $\widetilde{D}_{\theta_i}: p \in \mathbb{R}^3 \mapsto (\Delta p_i, \Delta s_i) \in \mathbb{R}^{3+1}$ which is conditioned by instance-specific neural weights $\theta_i$ and models a point deformation field $\Delta p$ and a SDF correction field $\Delta s$.
Instance-specific SDF field $s'_i$ can henceforth be expressed as $s'_i = T(p + \Delta p_i) + \Delta s_i$.
Borrowing from DIF meta-learning \cite{sitzmann2020implicit}, the instance-specific weights $\theta_i$ are obtained from a hyper-network $\Psi: \alpha_i \in \mathbb{R}^k \mapsto \theta_i \in \mathbb{R}^{|\Theta|}$.
This elegant formulation by DIF-Net authors enables the joint optimization of the models and instance codes $\alpha_i$, as well as dense correspondence for the modeled shape category. This solution was thus adopted in various subsequent studies \cite{alldieck2021imghum,zhang2022implicit}.

In this work, each sub-function also relies on its own deformation/correction field $D_j$ and hyper-net $\Psi_j$, though we propose a more comprehensive formulation of the deformation field, generalized from \cite{zhang2022implicit}. 
In order to jointly model several non-rigid objects, one not only needs to model intra-shape deformations, but also positional/scaling changes in objects across various instances.
Tackling dental geometry modeling, authors of DMM \cite{zhang2022implicit} made relatively simple assumptions about the transformation of their target classes (\texttt{gum} and \texttt{teeth}) and proposed specific deformation fields for each, to account for their different \textit{rigidness}. 
Here, we assume that per-object positional information is not provided and that per-category rigidness is not known. Therefore, we define our generalized deformation/correction function as:
\begin{equation}
D_{\theta_{i,j}}: p \in \mathbb{R}^3 \mapsto (r_{i,j}, t_{i,j}, \Delta p_{i,j}, \Delta s_{i,j}) \in \mathbb{R}^{3 + 3 + 3 + 1},  
\end{equation}
with the deformation explicitly composed of a rigid transformation $e^{\mathcal{S}_{i,j}}$ defined by the screw-axis $(r_{i,j}; t_{i,j})$ \cite{park2021nerfies} (\cf formula by \citealp{rodrigues1815attraction}) and non-rigid shape transformation $\Delta p_{i,j}$.
Hence, the $j$th SDF value of a point $p$ is expressed as:
\begin{equation}
    s'_{i,j} = T(p_{i,j}) + \Delta s_{i,j} \; \text{ with } \; p_{i,j} = e^{\mathcal{S}_{i,j}}p + \Delta p_{i,j}.
\end{equation}



\paragraph{Cross-Category Refinement.}
Each sub-function receives only local, object-specific information; so their correction field $\Delta s_{i,j}$ cannot account for inter-object relations. This can result in reconstructions with intersecting shapes and disregard for contact regions.
To mitigate this issue, we introduce a cross-category correction field $\Delta s_{\mathcal{O}_i}$ modeled by a final network $U$ based on concatenated signals from every sub-function. 
This correction vector is added to the $m$ SDF values as the final output.
In our implementation, each signal $\gamma_{i,j}$ originates from the penultimate activation of $D_j$, and $U$ is a shallow MLP with sine activation. Figure \ref{fig:ablation_study} highlights the significant impact of proposed $U$ on contact regions.

\subsection{Supervision}\label{sec:supervision}
Similar to previous DIF solutions, the proposed solution undergoes two optimization phases. First, the networks are optimized on a training dataset, along with the latent codes corresponding to each training instance (these codes can be discarded after training). Once trained, the pipeline can be used to reconstruct or annotate (via point correspondence) new instances. For each unseen instance, its conditioning code $\alpha_i$ is predicted via a shorter optimization phase, with the functions' weights frozen.

\paragraph{Sub-function Losses.}
Within each sub-function, we adopt the losses from DIF-Net \cite{deng2021deformed} to supervise SDF prediction (L1 accuracy of predicted SDF, correctness of surface normals, enforcing of Eikonal gradient property, cross-instance normal consistency, \etc), deformation (smoothness prior), and correction (regularization). We invite readers to access \cite{deng2021deformed} for details. We also apply L2 regularization to the latent codes, as suggested in \cite{park2019deepsdf}.
Additionally, we adapt the centroid loss from \cite{zhang2022implicit} to enforce that the centroid $c_{i,j}$ of shape $\mathcal{O}_{i,j}$ computed from post-deformation predictions coincides with the average centroid $\overline{c_j}$ estimated over the training samples. Based on our deformation formulation, the modified centroid loss is:
\begin{equation}
 \mathcal{L}_{j}^{centroid}= \Vert e^{\mathcal{S}_{i,j}}c_{i,j} + \Delta c_{i,j} - \overline{c_j} \Vert.
\end{equation}

\paragraph{Refinement Losses.}
We supervise the final SDF predictions $s_i$ using the curriculum SDF loss from \cite{duan2020curriculum}. Since the pre-refinement SDF fields are already supervised by the above-mentioned sub-function losses, we use the strictest tolerance and control parameters for the curriculum loss, \ie, $\varepsilon = 0$ and $\lambda = 0.5$.

Similar to the intra-category correction field, we also regularize the cross-category correction: 
\begin{equation}
    \mathcal{L}^{refreg} = \sum_{p \in \Omega} |\Delta s_{\mathcal{O}_i}(p)|.
\end{equation}





Finally, we introduce an attraction-repulsion supervision, \ie, contact loss, for off-surface points that lie in contact regions between two or more objects. 
\Ie, we define the set of these \textit{contact points} in training data as: 
$\mathcal{C} = \left\{(p, \Gamma) \mid \widehat{s}_{i,j} <\epsilon_c, j \in \Gamma , |\Gamma|\ge 2\right\}$, with $\Gamma$ the set of objects that $p$ is close to, $\widehat{s}_{i,j}$ the ground-truth SDF, and $\epsilon_c$ a hyper-parameter threshold (the smaller $\epsilon_c$, the thinner the considered contact regions). 
During optimization, to narrow the boundaries between these surfaces accordingly as well as avoid inter-penetration, we compute the following loss:
\begin{equation}
    \mathcal{L}^{\mathcal{C}} = \sum_{(p, \Gamma) \in \mathcal{C}} \sum_{j \in \Gamma}\sigma(|s'_{i,j}|)
\end{equation}
with $\sigma(s) = 2 \sigmoid(\lambda^{\mathcal{C}} s) - 1$, and $\lambda^{\mathcal{C}}$ weight controlling the constraint to the output SDF. 



The overall optimization objective is defined by linearly combining all the aforementioned losses, applying phase-specific loss weighting (see implementation details).

\section{Experiments}

\subsection{Experimental Protocol}

\paragraph{Implementation.}
For a fair comparison, we use the same MLP architectures as in DIF-Net \cite{deng2021deformed} for networks $\Psi_j$, $D_j$, $T_j$; only editing $D_j$ to return feature vector $\gamma_j$ along with its predictions.
We fix the dimensionality of per-object codes $\alpha_{i,j}$ to 128, and the value of $\lambda^{\mathcal{C}}$ to $10^2$. Other hyper-parameters are listed in annex.
Our model is trained on three NVIDIA RTX A40 GPUs for 300 epochs (\raisebox{-0.7ex}{\textasciitilde}1.5 hours over the WORD dataset).

\paragraph{Datasets.}
Since our model is focused on modeling sets of non-rigid objects, we opt for three medical shape benchmarks: WORD \cite{luo2022word}, AbdomenCT \cite{Ma-2021-AbdomenCT-1K} and Multi-Modality Whole-Heart Segmentation (MMWHS) \cite{zhuang2016multi}.
For WORD and AbdomenCT, we perform our evaluation on the following $m=6$ organs: \texttt{left-kidney}, \texttt{right-kidney} \texttt{liver}, \texttt{stomach}, \texttt{spleen}, and \texttt{pancreas}; noting the significant shape variability of some of these classes (\eg, \texttt{stomach}). 
For MMWHS, we consider 4 classes: \texttt{right-atrium}, \texttt{left-atrium}, \texttt{right-ventricle}, and \texttt{left-ventricle+left-myocardium} (merged into one shape as the left ventricle is contained inside the myocardium).
Each dataset contains the following number of samples: 30 training / 10 testing samples for MMWHS, 100/20 for WORD, and 37/10 for AbdomenCT (after removing 3 cases with livers cropped during scanning).


To generate training data points, we follow the sampling strategy proposed in \cite{zhang2022implicit}. \Ie, for each instance, we sample 200,000 surface points and 250,000 points randomly picked from $\Omega$ (normalized to [-1, 1]). 

\paragraph{Comparison.}
We compare to multiple state-of-the-art DIF methods: DeepSDF \cite{park2019deepsdf}, DIT \cite{zheng2021deep}, DIF-Net \cite{zhang2022implicit}, NDF \cite{sun2022topology}, and DMM \cite{zhang2022implicit}. 
Also composed of multiple DIF-Net submodules, DMM is the closest to our proposed model. However, it does not account for class interrelations and its deformation functions are tailored to specific classes (\texttt{gum} and \texttt{teeth}). For fair comparison, we create a custom version ``DMM (ada)'' of this pipeline that borrows our proposed deformation formulation.

\paragraph{Metrics.}
Like prior work \cite{park2019deepsdf}, we evaluate the shape reconstruction in terms of Chamfer distance (CD) and earth-mover distance (EMD). 
We also introduce an intersection volume (IV) metric to measure undesired interpenetration, as the volume (operator $\volume$) of the union of all pairwise 3D shape intersections in an instance:
\begin{equation}
    \IV_i = \volume(\mathcal{O}_{\cap}) \; \text{ with } \; \mathcal{O}_{\cap} = \underset{{(a, b) \in {m \choose 2}}}{\cup}\left(\mathcal{O}_{i,a} \cap  \mathcal{O}_{i,b}\right).
\end{equation}

    

\begin{table}[t]
  \centering
  \caption{Comparison to object-level reconstruction methods on 3 datasets (IV results are in kilo-units). Our method reliably outperforms other solutions, especially in terms of non-interpenetration.}
\resizebox{\linewidth}{!}{

    \begin{tabular}{@{\hskip0pt}l@{\hskip11pt}l|r@{\hskip3pt}c@{\hskip3pt}rr|r@{\hskip3pt}c@{\hskip3pt}rr|r}
    \toprule
    & \multirow{2}{*}{Models} & \multicolumn{4}{c|}{CD $\downarrow$} &\multicolumn{4}{c|}{EMD $\downarrow$} & IV $\downarrow$ \\ 
     & & \multicolumn{1}{c}{mean} & / & \multicolumn{1}{c}{std} & \multicolumn{1}{c|}{med.} 
     & \multicolumn{1}{c}{mean} & / & \multicolumn{1}{c}{std} & \multicolumn{1}{c|}{med.} & mean  \\
     
    \midrule

     \parbox[t]{0mm}{\multirow{5}{*}{\rotatebox[origin=c]{90}{\textbf{WORD}}}}
     & \footnotesize{DeepS.} & \underline{16.68} & / & \underline{5.58}& 17.23  & \underline{8.24} & / & \underline{0.83} & 8.29  & 11.39 \\
    & DIT & 25.10 & / & 8.95 & 22.46  & 8.33 & / & 1.01 & 8.13  & 39.08 \\
    & DIF & 19.01 & / & 7.11 & 17.01  & 8.50 & / & 1.04 & 8.29  & 16.66 \\
    & NDF & 24.28 & / & 63.79 & \textbf{8.46} & 8.72 & / & 3.21 & \textbf{7.91} & \underline{8.96} \\
    \cmidrule{2-11}
    & \textbf{Ours} & \textbf{14.63} & / & \textbf{3.73} & \underline{14.36}  & \textbf{8.01} & / & \textbf{0.83} & \underline{8.00}  & \textbf{4.24} \\
    
    \midrule
    \midrule
    
    \parbox[t]{0mm}{\multirow{5}{*}{\rotatebox[origin=c]{90}{\textbf{MMWHS}}}}
     & \footnotesize{DeepS.}  & 7.87 & / & \underline{3.82} & 7.17  & 7.31 & / & 2.00 & 7.10  & 19.79 \\
    & DIT & 23.95 & / & 30.57 & 11.98  & 8.23 & / & 2.58 & 7.94  & 30.29 \\
    & DIF & 11.37 & / & 6.37 & 10.68  & 7.63 & / & 2.50 & \underline{6.98}  & \underline{6.22} \\
    & NDF & \textbf{4.37} & / & 6.83 & \textbf{2.34} & \underline{7.09}& / & \textbf{1.87} & 7.38  & 7.05\\
    \cmidrule{2-11}
    & \textbf{Ours} & \underline{4.95} & / & \textbf{1.67} & \underline{4.83}& \textbf{6.80} & / &  \underline{1.98} & \textbf{6.88} & \textbf{1.52}\\
    
    \midrule
    \midrule
    
    \parbox[t]{0mm}{\multirow{5}{*}{\rotatebox[origin=c]{90}{\textbf{AbdomenCT}}}}
     & \footnotesize{DeepS.}   &\underline{41.65} & / & \underline{12.33} & \textbf{37.06 }& 11.13 & / & 1.10 & 11.08 &20.09\\
    & DIT & 66.27 & / & 13.92&65.74& 12.14 & / & 1.26& 11.88 &109.35 \\
    & DIF& 45.67 & / & 15.78 & \underline{40.13} & \underline{10.90} & / & \underline{0.87}& \underline{10.90}&\underline{16.37} \\
    & NDF & 60.77 & / & 30.23 & 56.11 & 12.37 & / & 2.54& 11.67 & 56.02  \\
    \cmidrule{2-11}
   & \textbf{Ours} & \textbf{41.60} & / & \textbf{9.31} & 41.08&\textbf{10.70} & / & \textbf{0.59} & {\textbf{10.81}} & \textbf{3.40} \\
    \bottomrule
    \end{tabular}%
}
  \label{tab:table_results}%
\end{table}%

\begin{table}[t]
  \centering
  \caption{Comparison to instance-level reconstruction methods on merged instance meshes from WORD dataset.}

\resizebox{\linewidth}{!}{
    \begin{tabular}{l|r@{\hskip3pt}c@{\hskip3pt}rr|r@{\hskip3pt}c@{\hskip3pt}rr}
    \toprule
    \multirow{2}{*}{Methods}&\multicolumn{4}{c|}{CD $\downarrow$} &\multicolumn{4}{c}{EMD $\downarrow$}\\
     & \multicolumn{1}{c}{mean} & / & \multicolumn{1}{c}{std} & \multicolumn{1}{c|}{median} 
     & \multicolumn{1}{c}{mean} & / & \multicolumn{1}{c}{std} & \multicolumn{1}{c}{median} \\
    \midrule
    DeepSDF & 108.75 & / & 29.97  & 103.68  & 15.96  & / & 1.96  & 15.62  \\
    DIT & 303.44 & / & 45.04  & 297.43  & 20.15 & / & 2.64  & 19.45  \\
    DIF & 72.96& / &21.47  & 72.83  &14.52 & / &  2.50  & 14.57  \\
    NDF & 128.35 & / & 78.37  & 91.92  & 18.18 & / & 5.47  & 17.18  \\
    DMM {\tiny(\texttt{teeth})} & 205.71 & / & 66.57  & 195.78&17.76 & / & 3.88  & 16.96  \\
    DMM {\scriptsize(\texttt{gum})} & 215.40 & / & 70.08  & 196.82  & 18.19 & / &  3.25  & 17.17  \\
    DMM {\scriptsize(ada)}&\underline{23.62} & / & \underline{7.03}& \underline{23.12}& \underline{10.10} & / & \underline{1.15}&\underline{10.19}\\ 
    
    \midrule
    \textbf{Ours} & \textbf{14.63} & / & \textbf{3.73} & \textbf{14.36} & \textbf{8.01} & / & \textbf{0.83} & \textbf{8.00} \\
    \bottomrule
    \end{tabular}%
    }
  \label{tab:word_results}%
\end{table}%

\begin{figure*}[t]
\centering
    \includegraphics[width=1.\textwidth]{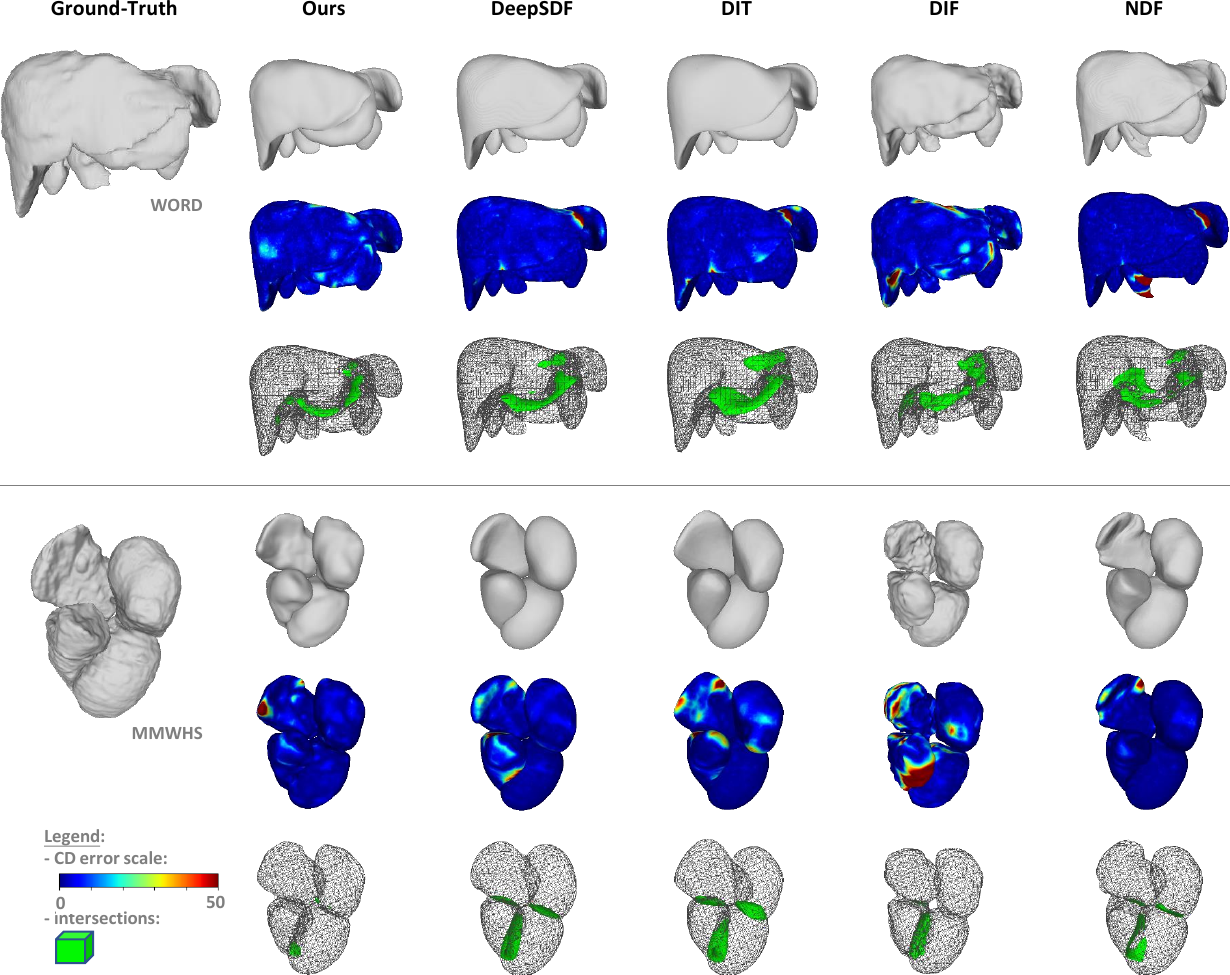}
    \caption{Qualitative comparison to prior art, plotting the reconstruction accuracy (CD-based color coding) and the erroneous intersections (green-colored regions) to highlight the positive impact of proposed inter-objects signals.}
    \label{fig:reconstruct}
\end{figure*}

\subsection{Shape Representativeness}
To evaluate representation capability, we challenge the methods to reconstruct instances not seen during training; by fixing the methods' weights, optimizing the latent code corresponding to the new shape(s), and finally measuring the quality of the reconstructed instance (\cf Supervision subsection). 
Since prior art focuses on outputting a single SDF, we propose two different settings for other methods: \textit{object-level} and \textit{instance-level} reconstructions. 
\paragraph{Object-Level Reconstruction.}
In this experiment, we train $m$ different instances of each prior method (except DMM which can generate segmented entities), each separately modeling one category. We thus make the assumption that the centroids and scales of each object are available to these models during inference, which is unfair to our method that learns by itself these object transformations.

\paragraph{Instance-Level Reconstruction.} 
In this scenario, we merge the $m$ shapes of each instance into one, and train prior methods to model this global geometry. Compared to ours, these solutions under-perform, losing category information (IV metric thus cannot be computed) and having trouble modeling contact regions.

    

\paragraph{Results.}
Tables \ref{tab:table_results} and Table \ref{tab:word_results} show that our model achieves the highest reconstruction accuracy on all datasets, despite solving a higher complexity task compared to object- and instance-level prior methods. In comparison, NDF \cite{sun2022topology} can achieve accurate results on objects with stable topological features but occasionally fails for instances with larger deformations.  
IV results also show that our method effectively reduces cross-object inconsistencies without sacrificing reconstruction accuracy. 
Qualitative examples are provided in Figure \ref{fig:reconstruct} and supplementary material.

\subsection{Ablation Study}
We confirm the significance of our technical contributions (generalized deformation, cross-category shape correction, contact loss) via an ablation study on the WORD dataset \cite{luo2022word}.
Figure \ref{fig:ablation_study} also provides further insight into the impact of the different algorithmic steps on the predicted shapes. \Eg, in the provided sample, we can observe how the \texttt{liver}/\texttt{stomach} and \texttt{liver}/\texttt{left-kidney} contact regions are improved by our refinement function.

\begin{table}[t]
  \centering
  \caption{Ablation study on WORD dataset, evaluating our work \textit{without} key contributions (more results in annex).}

\resizebox{\linewidth}{!}{
    \begin{tabular}{r@{\hskip5pt}rrrrrr}
    \toprule
    Without: & 
    \multicolumn{1}{c}{$\mathcal{L}^{\mathcal{C}}$ \& $\Delta s_{\mathcal{O}}$} & 
    \multicolumn{1}{c}{$\mathcal{L}^{\mathcal{C}}$} & 
    \multicolumn{1}{c}{$\Delta s_{\mathcal{O}}$} & 
    \multicolumn{1}{c}{$\Delta p_j$} & 
    \multicolumn{1}{c}{$(r_i; t_i)$} & 
    \multicolumn{1}{c}{$\O$} \\
    \midrule
    CD $\downarrow$ & 15.90 & 16.36 & 15.33 & 15.06 & 15.44 & \textbf{14.63}\\
    EMD $\downarrow$ & 8.43 & 8.67 & 9.17 & 8.10 & 8.27 & \textbf{8.00}\\
    \bottomrule
    \end{tabular}%
    }
  \label{tab:ablation}%
\end{table}%

    



\subsection{Applications}

\begin{figure*}[htbp]
\centering
    \includegraphics[width=1.\textwidth]{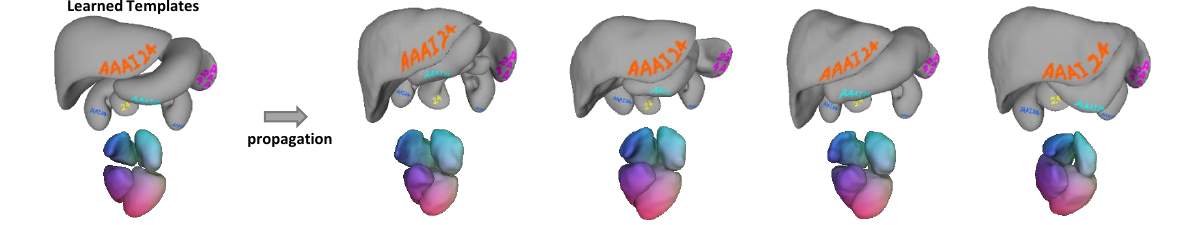}
    \caption{
    Visualization of propagating annotations by leveraging dense correspondences among reconstructions. The manual surface annotations on the learned template (left) are propagated to various instances (right). 
    }
    \label{fig:point correspondence}
\end{figure*}

\paragraph{Point Correspondences.}
We showcase the point correspondence capability of MODIF in Figure \ref{fig:point correspondence} and in annex. 
While the mechanisms enabling accurate correspondences are borrowed from the literature \cite{deng2021deformed}, our method is the only one to simultaneously provide separate per-category template meshes and point correspondences. 
This means, for instance, that our method can not only accurately extrapolate the shape of any object missing from a new instance (as shown in the next experiment), but it can also accurately transfer any dense annotation to predicted surfaces. We believe that this may have significant applications, \eg, in clinical data annotation and analysis.

\begin{table}[t]
  \centering
  \caption{Comparison to per-object NDF on the missing-object recovery task (missing \texttt{left-myocardium} in MMWHS test instances). We separately report metrics computed over the recovered missing organs (``miss.") and over the reconstructed present organs (``pres.").}
\resizebox{\linewidth}{!}{

    \begin{tabular}{@{\hskip0pt}l@{\hskip11pt}l|r@{\hskip3pt}c@{\hskip3pt}rr|r@{\hskip3pt}c@{\hskip3pt}rr|r}
    \toprule
    & \multirow{2}{*}{Models} & \multicolumn{4}{c|}{CD $\downarrow$} &\multicolumn{4}{c|}{EMD $\downarrow$} & IV $\downarrow$ \\ 
     & & \multicolumn{1}{c}{mean} & / & \multicolumn{1}{c}{std} & \multicolumn{1}{c|}{med.} 
     & \multicolumn{1}{c}{mean} & / & \multicolumn{1}{c}{std} & \multicolumn{1}{c|}{med.} & mean  \\
     
    \midrule

     \parbox[t]{0mm}{\multirow{3}{*}{\rotatebox[origin=c]{90}{\textbf{miss.}}}}
     & NDF & 142.19 & / &  82.29  & 125.92  & 11.47 & / & 3.01 & 11.12 & 33.69  \\
     & DIF & 94.44 & / &  54.44  & 79.32  & 10.84 & / & 3.20  & 10.28 & 16.80 \\
      & \textbf{Ours} & \textbf{45.52} & / & \textbf{27.06} & \textbf{37.74} & \textbf{6.46} & / & \textbf{2.09} & \textbf{5.73} & \textbf{1.20}  \\
      
    \midrule
    \midrule
    
     \parbox[t]{0mm}{\multirow{3}{*}{\rotatebox[origin=c]{90}{\textbf{pres.}}}}
     & NDF & 6.06 & / & 18.88  & \textbf{2.19} & \textbf{2.81} & / & 1.87  & \textbf{2.39} & 35.73  \\
     & DIF & 13.65 & / & 10.07 & 12.19 & 3.96 & / & 1.42 & 3.58 & 12.36 \\
      & \textbf{Ours} & \textbf{5.77} & / & \textbf{3.01} & 5.33 & 3.11 & / & \textbf{0.61} & 3.07  & \textbf{1.47} \\
    
    \bottomrule
    \end{tabular}%
}
  \label{tab:missing-organ}%
\end{table}%

\paragraph{Missing-Object Recovery.}
Our method uniquely combines per-category shape prediction (via its sub-functions $f_j$, performing similarly as prior object-level methods) and cross-category relation modeling (\cf refinement $U$).
These properties make it possible for MODIF to tackle an under-studied task: missing-object recovery, \ie, when an instance is missing one of its objects (\eg, organ not properly captured during scanning or not properly segmented by the annotator).

In such cases, our solution can rely on its knowledge about the missing shape's overall distribution (\cf learned template), as well as information extracted from non-missing objects in the instance, to recover a statistically plausible shape, that is consistent with other present objects (matching scale/positioning, minimal overlap, plausible contact surfaces, \etc).

To test our method on this task, we consider the MMWHS dataset (more challenging due to the higher ratio of contact regions), removing the \texttt{left-myocardium} shape from all test instances. We compare to per-object NDF (best challenger on WORD) and DIF. We adapt NDF inference so that the model returns the template shape for the missing organ, with its scale and position adjusted according to the mean rigid transformation between the centroids of predicted non-missing shapes and the centroids of their corresponding templates. 

\begin{figure}[H]
    \centering
    \includegraphics[width=.92\linewidth]{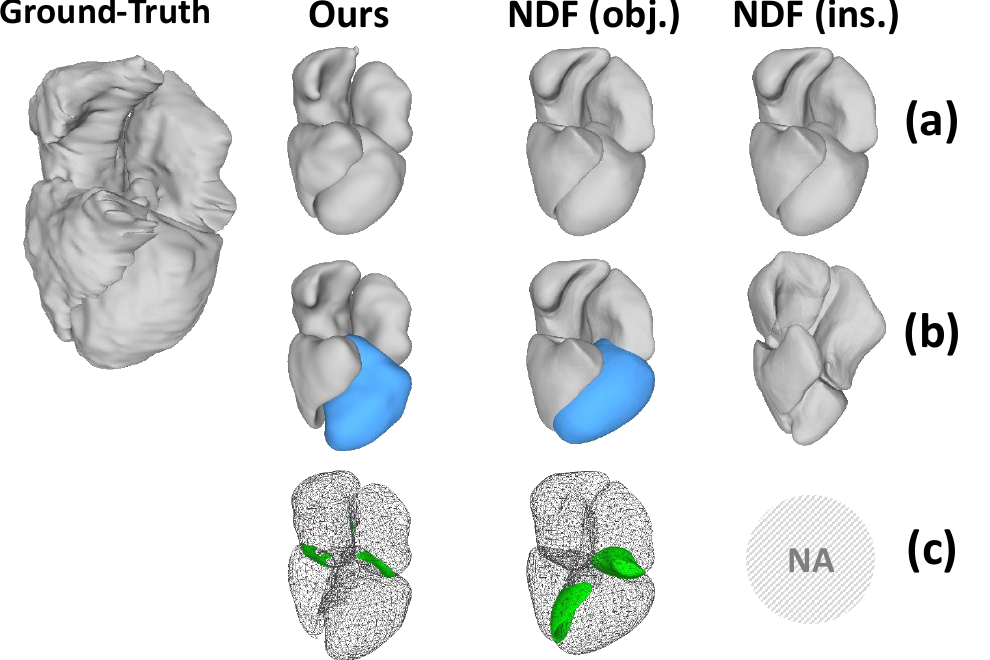}
    \caption{Qualitative results for the missing organ task.
    Row (a) shows reconstructed results when all organs are provided. Row (b) shows predictions when the left myocardium (blue shape) is missing from the instance. Row (c) shows erroneous overlaps in predicted organs.
    }
    \label{fig:missing_organ}
\end{figure}

Results are shared in Table \ref{tab:missing-organ}, demonstrating the superiority of our method, which successfully optimizes the recovered shape to fit with the other organs (both in terms of rigid and soft transformation). The table also shows that this recovery does not negatively impact the reconstruction of other present shapes. Note that we also try to apply the instance-level NDF to this recovery task but observed worse results (mean CD = 93.41, mean EMD = 12.63). We provide a qualitative comparison in Figure \ref{fig:missing_organ}, highlighting the minimized interpenetration.

\begin{figure}[t]
    \centering
    \includegraphics[width=1\linewidth]{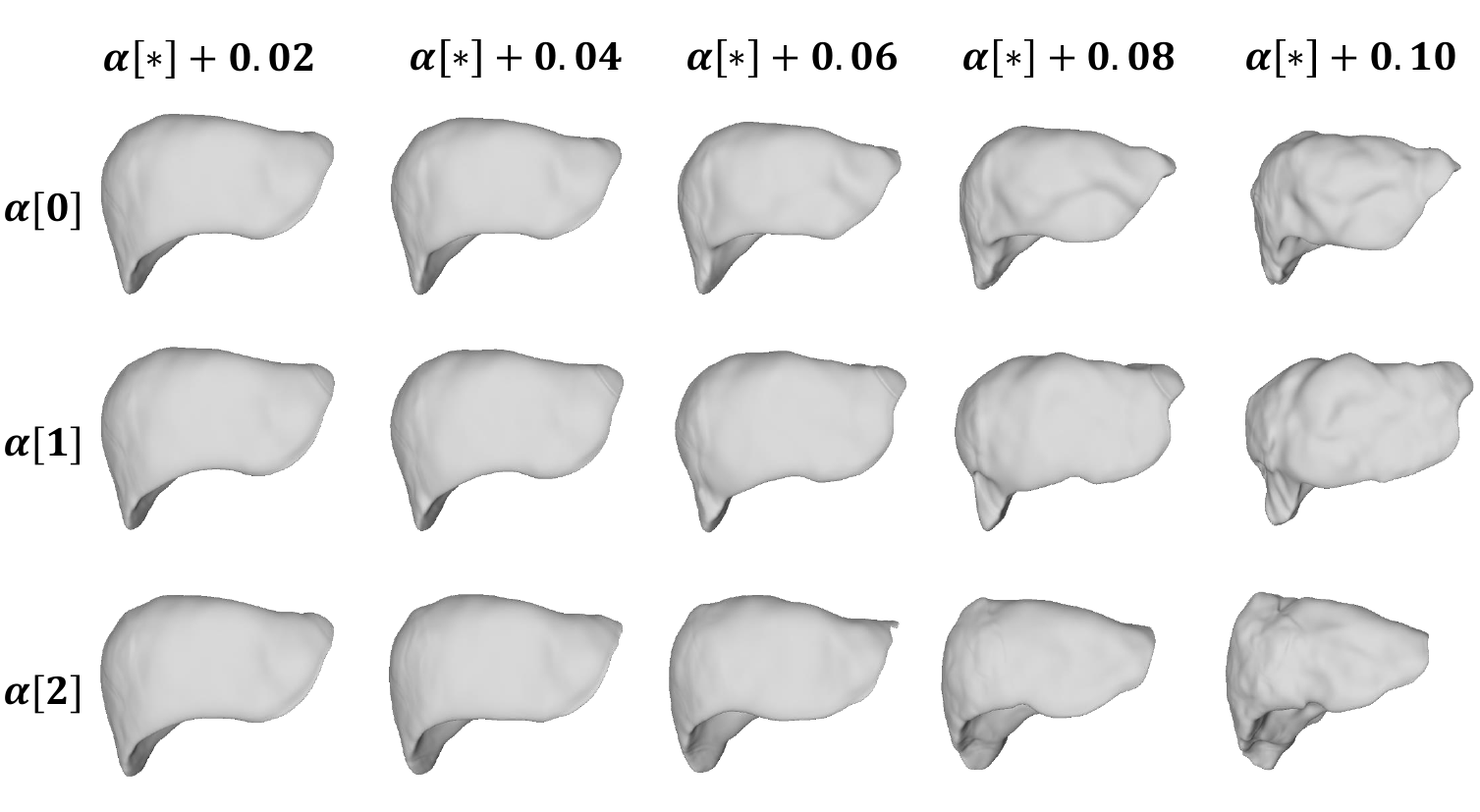}
    \caption{Realistic data augmentation by latent code editing.}
    \label{fig:latent_change}
\end{figure}

\paragraph{Data Augmentation.}
By learning the general data distribution, our method can generate realistic anatomical digital twins for dataset augmentation, training, \etc. By tweaking learned latent code values, we are able to generate new synthetic instances, as shown in Figure \ref{fig:latent_change}.


\section{Conclusion}
We proposed MODIF, a novel implicit neural function to model multi-object 3D instances. Iterating over previous single-category SDF models, our solution relies on a cross-category refinement mechanism and a contact loss to model the correlation between objects. 
Experiments on various datasets show that our model performs better than prior art in terms of reconstruction accuracy and is highly effective at reducing interpenetration problems. 
Furthermore, we introduce the task of missing-object recovery from incomplete instance set and demonstrate that our model can predict accurate and consistent shape compared to other methods, thanks to our cross-category signaling. 


\paragraph{Limitations.} It should be noted that, similar to previous methods, our model expects class-labeled shapes, which can be a limitation for some scenarios (\eg, non-classified instances, open-set instances, \etc). Our deformation function may also be considered overly generic. Prior knowledge on specific categories could be injected to better guide the model (\eg, borrowing from the extensive literature on non-penetrating non-rigid simulation \cite{baraff1993non}).

\paragraph{Societal Impact.} 
The above limitation may constrain the adoption of our method to specific use cases, and the statistical modeling of anatomical data may penalize patients with unique conditions (\textit{outliers}). On the other hand, we would argue that the benefits in terms of automated dense data annotation outshine these limitations, with proper safeguards (\eg, validation by experts). 
All in all, we believe that the proposed method can positively impact both the computer-vision and medical communities, by offering a novel model for multi-SDF learning and demonstrating its values on medical benchmarks.

\section{Acknowledgments}
This work is supported by the National Key R\&D Program of China (2021ZD0111100) and sponsored by Shanghai Rising-Star Program.

\bibliography{aaai24}

\appendix
\beginsupplement
\newpage
\noindent
\begin{center}
\pdfbookmark[0]{Supplementary Material}{sup_mat}
{\Large\textbf{\centering \MakeUppercase{Supplementary Material}} \vspace{1em}}
\end{center}
In this supplementary material, we provide further methodological context and implementation details to facilitate reproducibility of our framework MODIF. We also showcase additional quantitative and qualitative results to further highlight the contributions claimed in the paper. 

\section{Implementation Details}
\subsection{Network Architectures}

\paragraph{Sub-function $\bm{f_j}$.}
For each of the three networks composing the sub-functions (all MLPs with sine activations, as proposed by \cite{sitzmann2020implicit}), we borrow their architecture from the official implementation of \cite{zhang2022implicit}, based on \cite{deng2021deformed}.




We fix the size of the per-category latent codes $\alpha_{i,j}$ (passed to the hyper-net $\Psi_j$) to 128.

We edit the neural deformation fields $D_j$ (with weights conditioned by  $\Psi_j$) so that this network outputs our deformation parameters $(r_{i,j}, t_{i,j}, \Delta p_{i,j}, \Delta s_{i,j}) \in \mathbb{R}^{10}$, as well as the category-specific feature vector $\gamma_{i,j} \in \mathbb{R}^{64}$. This vector corresponds to the features returned by the penultimate layer of the MLP.

\paragraph{Refinement Field $\bm{U}$.}
Our cross-category correction function $U: \mathbb{R}^{64m} \mapsto \mathbb{R}^{m}$ is a two-layer MLP, with an inner feature space of size $128$ and sine activations.

\subsection{Supervision}

\paragraph{Losses.}
In the following paragraphs, we detail the additional loss functions used to supervise our solution, that were not described in the main paper (\ie, please refer to the main paper for the new losses).

\vspace{.2em}
\noindent
\textit{Sub-function Losses. }
Each sub-module is mainly supervised by the following criterion, adapted from \cite{deng2021deformed}, to learn individual shapes (note that we drop the subscript $i$ from notations in the following equations, for clarity).  

\begin{equation}
\begin{aligned}
    \mathcal{L}_{j}^{sdf} = &\sum_{p \in \mathcal{O}_j}\left(|s'_j|+(1-\langle\nabla s'_j, \overline{n}_j\rangle)\right) \\&+ \sum _{p \in \Omega} |\Vert\nabla f_j(p)\Vert_{2} -1| + \sum_{p \in \Omega \backslash \mathcal{O}_j} \rho(s'_j),
 \end{aligned}
    \label{subsdf}
\end{equation}
where $\mathcal{O}_j$ is the surface of object $j$, $\Omega$ is the 3D space, and $\rho(s) = e^(-\delta \cdot |s|), \delta \gg 1$ is a function encouraging off-surface points to have an SDF value other than zero (last term).
Note that our first term is partially different from \cite{deng2021deformed}.
We also enforce the gradients of predicted SDF values for on-surface points to be equal to ground-truth surface normals $\overline{n}_j$. However, we relax the need for off-surface ground-truth SDF values by considering only on-surface points.
The second term enforces the Eikonal equation, \ie, that the norm of the spatial gradient should be 1.

As in \cite{park2019deepsdf}, we use a regularization loss to constrain the learned latent code:
\begin{equation}
    \mathcal{L}_{j}^{reg} = \Vert \alpha_{j} \Vert_{2}^2
\end{equation}

Intra-object correction fields also regularized by minimizing the term:
\begin{equation}
    \mathcal{L}_{j}^{correg} = \sum_{p \in \Omega} |\Delta s_{}|
\label{correction1}
\end{equation}

To generate consistent correspondences for objects having diverse shapes, we constrain the deformation field by enforcing normal consistency between points in the template space and their correspondences in all given instances \cite{deng2021deformed}. 

 \begin{equation}
     \mathcal{L}_{j}^{normal} = \sum_{p \in \mathcal{O}_j} (1-\langle\nabla T_j(p_{j}), \overline{n}\rangle) 
\end{equation}
where $\nabla T$ is the spatial gradient of the template field $T$. 

We also implement the deformation smoothness prior of \cite{deng2021deformed} by penalizing its spatial gradient. 

\begin{equation}
    \mathcal{L}_{j}^{smooth} = \sum_{p \in \Omega} \Vert \nabla(p_{j} - p) \Vert_{2}.
\end{equation}

\vspace{.2em}
\noindent
\textit{Refinement Losses}

We use the curriculum SDF loss defined in \cite{duan2020curriculum}. The loss has two parameters: a tolerance parameter $\varepsilon$ and a control parameter $\lambda$ to control the significance of hard and semi-hard samples.
\begin{equation}
    \mathcal{L}^{sdf} = \mathcal{L}_{\varepsilon, \lambda}(f(\alpha, p), \widehat{s})
\end{equation}
Considering each sub-module is already being supervised by Equation \ref{subsdf}, we use the strictest requirement for the final SDF prediction, \ie, setting $\varepsilon = 0$ and $\lambda = 0.5$.

\paragraph{Per-Phase Objectives.}
As mentioned in the main paper, we consider two optimization phases: the training of the whole solution, and then the optimization of latent codes to fit new instances (with network weights frozen).  

\vspace{.2em}
\noindent
\textit{Training.}
Henceforth, the objective function during the training phase is formulated as :
\begin{equation}
\begin{aligned}
    \mathcal{L} = \frac{1}{m} \sum_{j=1}^{m} &\big(\lambda^{sdf}\mathcal{L}_j^{sdf} + \lambda^{reg}\mathcal{L}_j^{reg} + \lambda^{correg}\mathcal{L}_j^{correg}\\&+ \lambda^{normal}\mathcal{L}_j^{normal} +\lambda^{smooth} \mathcal{L}_j^{smooth}\\& + \lambda^{centroid}\mathcal{L}_j^{centroid}\big)
    \\ + \lambda^{fsdf}&\mathcal{L}^{sdf} + \lambda^{refreg}\mathcal{L}^{refreg} + \lambda_{\mathcal{C}}\mathcal{L}^{\mathcal{C}},
\end{aligned}
\label{eq:train_losses}
\end{equation}
where $\lambda^{\boldsymbol{\cdot}}$ are the weights applied to each loss term.

During training, we set the balancing weights as $\lambda^{sdf} = 1, \lambda^{reg} = 1e3, \lambda^{correg} = 5e2, \lambda^{normal} = 1e2, \lambda^{smooth} = 5, \lambda^{centroid} = 1, \lambda^{fsdf} = 5e2, \lambda^{refreg}= 3e2, \lambda^{\mathcal{C}} = 5$.

\vspace{.2em}
\noindent
\textit{Reconstruction.}
The reconstruction phase is aimed at optimizing only the latent codes to fit new instances, hence with function weights frozen. Therefore, we relax the regularization terms (except the one applied to latent codes $\alpha_j$) and normalization terms, to focus the optimization towards surface accuracy. 

The following loss weights are thus set to zero: $\lambda^{refreg} = \lambda^{correg} = \lambda^{normal} = \lambda^{smooth} =  \lambda^{centroid} = 0$.  The remaining weights are assigned the following values: $\lambda^{sdf} = 1, \lambda^{reg} = 1e3, \lambda^{fsdf} = 5e2, \lambda^{\mathcal{C}} = 5$.



\subsection{Data Preparation}
Every training and testing multi-shape instance $\mathcal{O}_i$ undergoes a data pre-processing phase, to convert its explicit surface information (\ie, $m$ lists of vertices and faces forming the shape meshes) into SDF supervision (\ie, on-surface and off-surface points with corresponding vectors of SDF values \wrt each of the $m$ objects; and corresponding surface normal vectors for on-surface points only).

For this pre-processing step, we follow the standard procedure adopted in prior works \cite{deng2021deformed,alldieck2021imghum,zhang2022implicit,sun2022topology}, with point sampling parameters borrowed from DIF \cite{deng2021deformed} and DMM \cite{zhang2022implicit}, but adapted to our multi-object scenarios.
\Ie, we normalize the ground-truth meshes of every instance $\mathcal{O}_i$ so that the whole instance / all meshes fit into a sphere with a radius of 1 (prior art would instead normalize each mesh separately).
Then, we use the \texttt{mesh-to-sdf} library \cite{kleineberg_calculate_2023} to randomly sample surface points (default seed is applied). 
The per-object number of surface points for each object $\mathcal{O}_{i,j}$ is decided by its surface area. Each point sampled \wrt surface $k$ comes with its SDF values $\widehat{s}_i = \{\widehat{s}_{i,j}\}_{j=0}^m$ and surface normal vector $\overline{n}_j$, where $\widehat{s}_{i,k} = 0$ and where $\widehat{s}_{i,j\neq k}$ are computed by the library. 
For free-space points, we uniformly sample them from a cube spawning $[-1.5,1.5]$ in every dimension. Note that we set a larger domain for the free points compared to the normalized instance dimensions, to make sure that the sampling covers the entire instance. 
For each free point, we calculate its distance to the nearest surface of each object and get the corresponding ground-truth SDF value $\widehat{s}_{i,j}$. 

In summary, for each instance,  we randomly sample 200K surface points with labels and normals, together with 250K free-space points with their SDF values to each object. 

\subsection{Training Parameters}

We initialize the per-instance per-object latent codes $\alpha_{i,j}$ by sampling from the normal distribution $\mathcal{N}(0,0.01^2)$ as \cite{park2019deepsdf}. The weights of template fields $T_j$, hyper-nets $\Psi_j$, and  deformation fields $D_j$ are all initialized following \cite{deng2021deformed}. Our refinement model $U$ is initialized according to \cite{sitzmann2020implicit}. 

For each benchmark, we train our model for 300 epochs using the Adam optimizer \cite{kingma2014adam} with a learning rate of $1e-3$ and batch size of 12 instances.  At each iteration, we randomly select 16,384 points for each instance in the batch and calculate the loss functions defined in Equation \ref{eq:train_losses}. For inference, the latent code is optimized for 800 iterations with a learning rate of $1e-2$.

For prior art (DeepSDF, DIT, DIF, NDF), we use their official implementations and default settings.

\subsection{Point Correspondence}
\begin{figure*}[t]
    \centering 
    \includegraphics[width=1\linewidth]{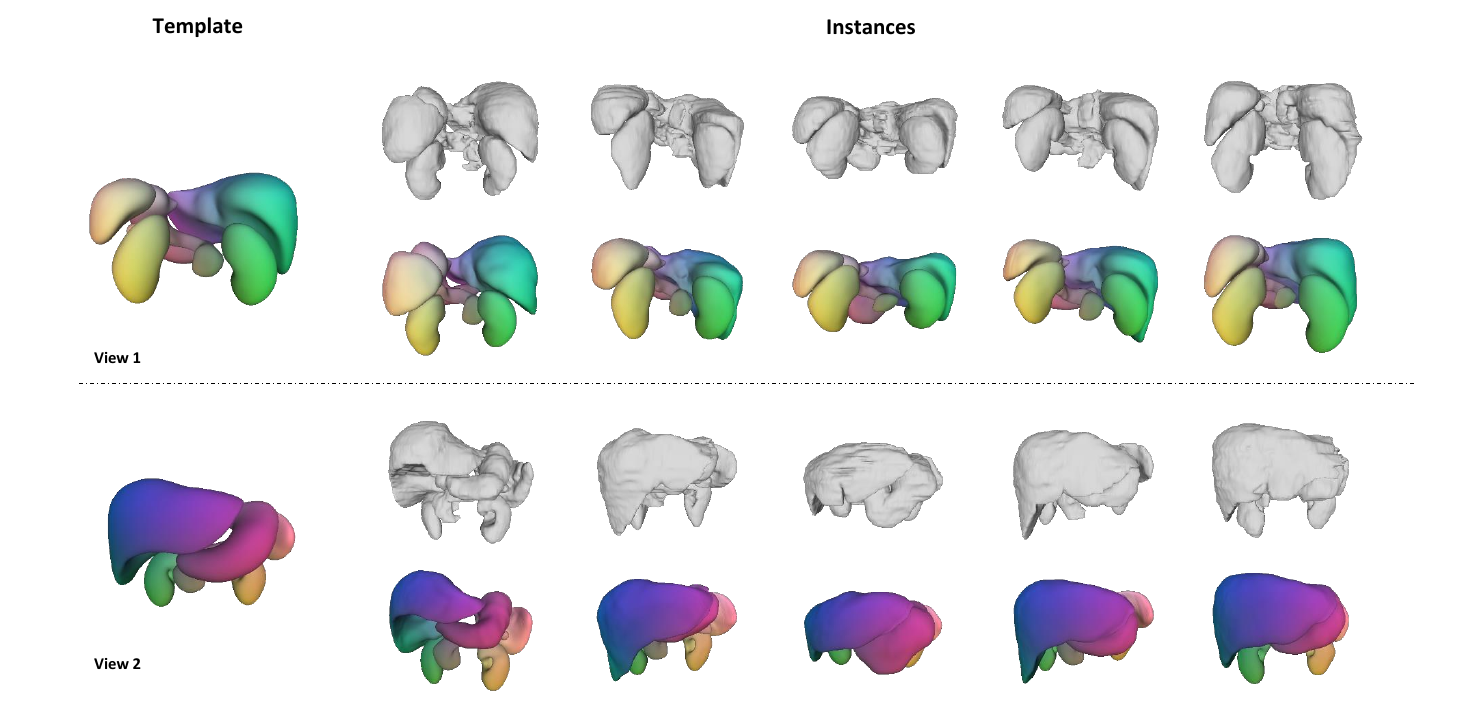}\\
    \caption{Visualization of cross-instance multi-object point correspondence on WORD data \cite{luo2022word}.
    }
    \label{fig:corres_word}
\end{figure*}

\begin{figure*}[!h]
    \centering 
    \includegraphics[width=1\linewidth]{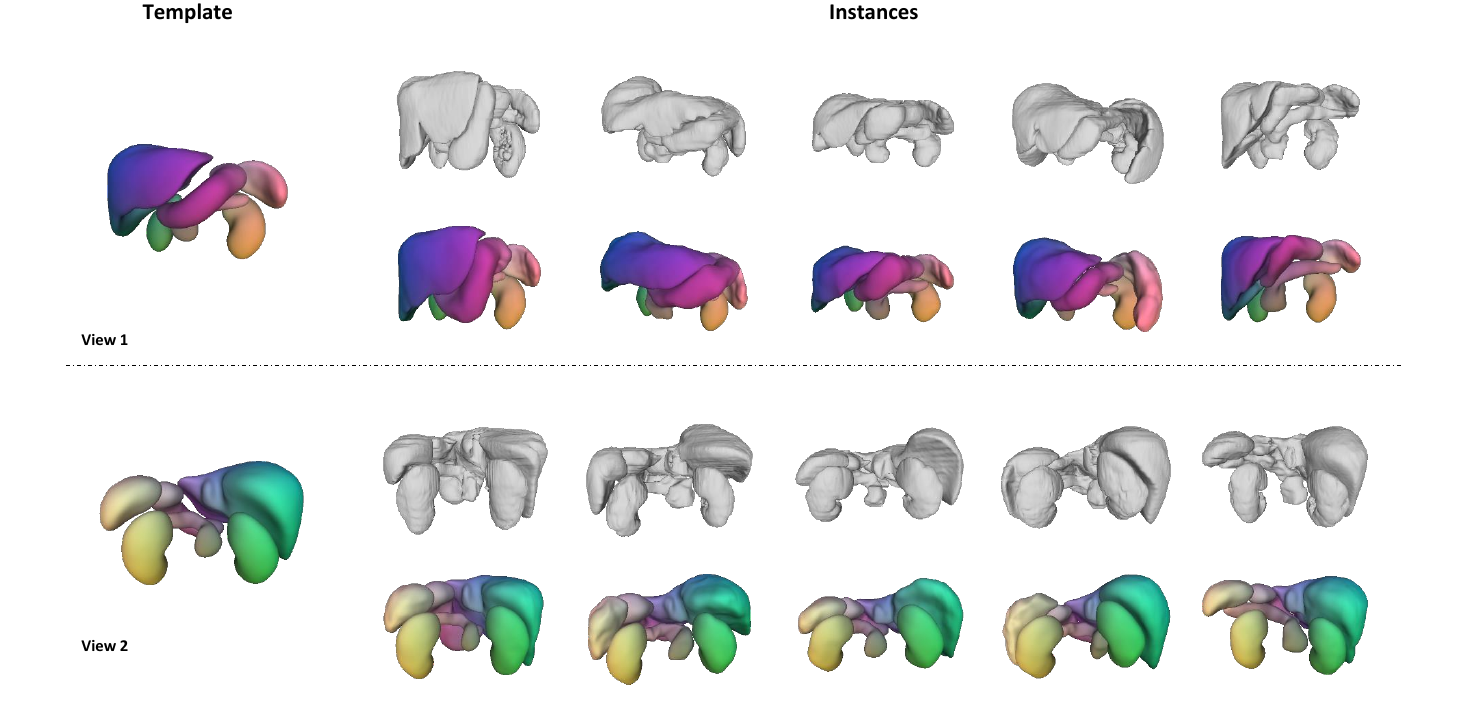}\\
    \caption{Visualization of cross-instance multi-object point correspondence on AbdomenCT-1K data \cite{Ma-2021-AbdomenCT-1K}.
    }
    \label{fig:corres_abdo}
\end{figure*}

\begin{figure*}[h]
    \centering 
    \includegraphics[width=1\linewidth]{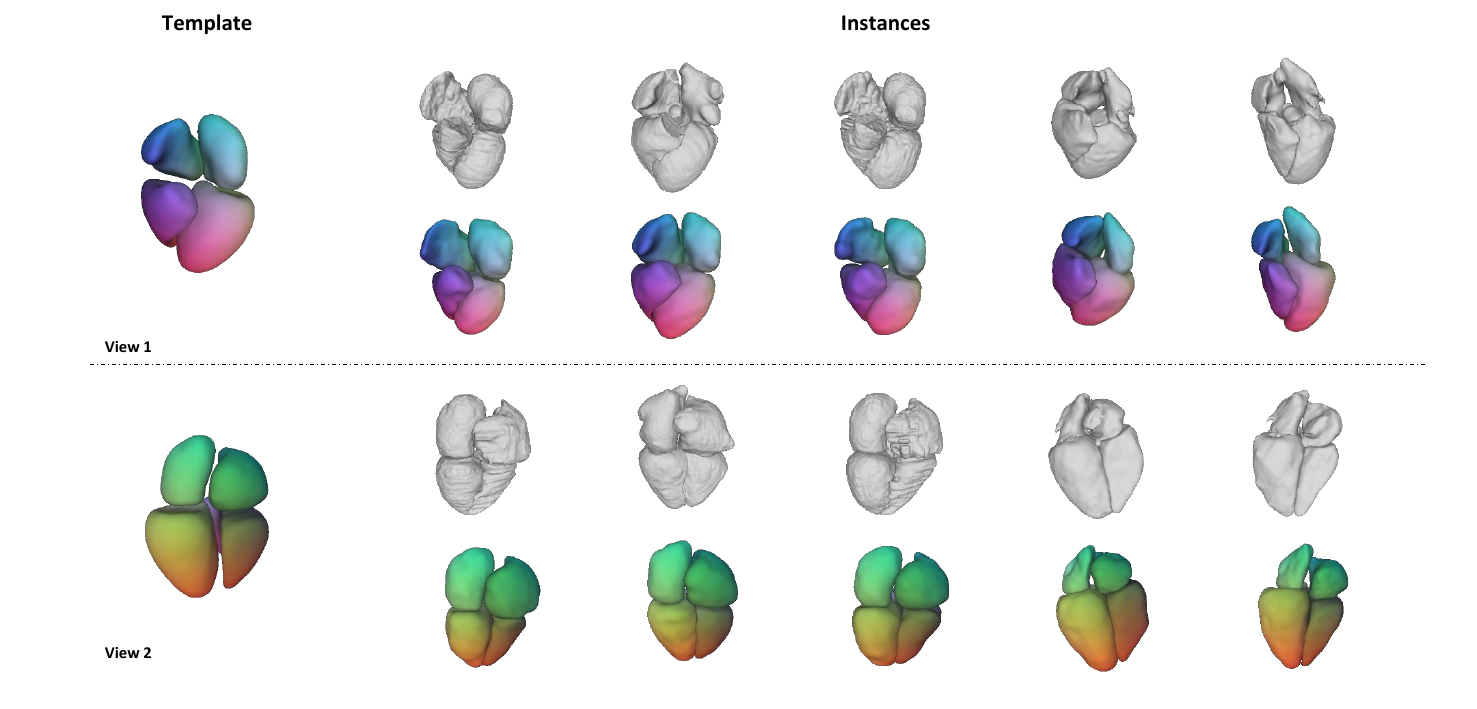}\\
    \caption{Visualization of cross-instance multi-object point correspondence on MMWHS data \cite{zhuang2016multi}.
    }
    \label{fig:corres_mmwhs}
\end{figure*}

In Figures \ref{fig:corres_word},\ref{fig:corres_abdo}, and \ref{fig:corres_mmwhs}, we provide additional qualitative results demonstrating the multi-object point correspondence capability of the proposed method. We can observe the consistency of the dense surface annotations (color gradient here) across instances, as well as with the corresponding multi-object template.

\subsection{Instance-level Reconstruction}
\begin{figure}
    \centering
    \includegraphics[width=.9\linewidth]{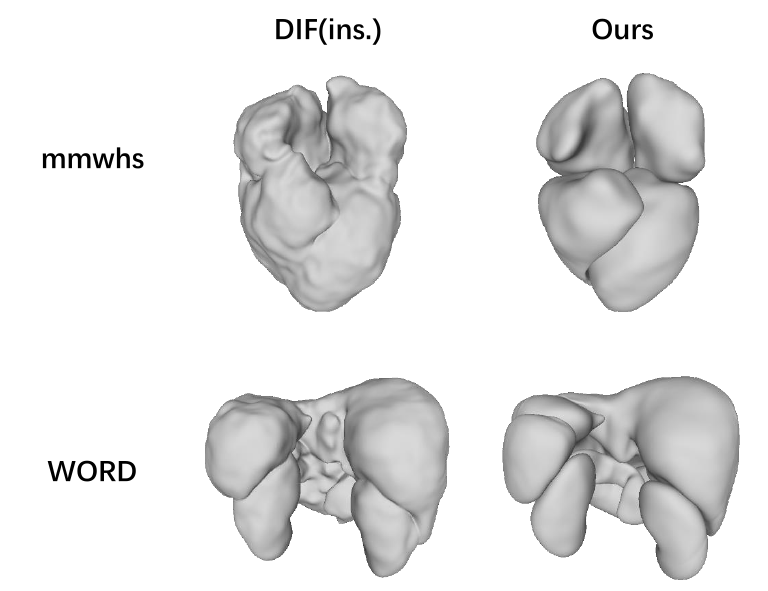}
    \caption{Prior work fails to model instance-level contact regions.}
    \label{fig:inst_reconstruct}
\end{figure}

Prior methods output a single SDF value per point. As a result, they suffer from heavy discontinuities of the instance-level distance field in contact regions. For example, Figure \ref{fig:inst_reconstruct} shows severe adhesion between organs for DIF-Net.

\section{Additional Experiments}

\subsection{Ablation Study}
\begin{figure*}[t]
    \centering 
    \includegraphics[width=1\linewidth]{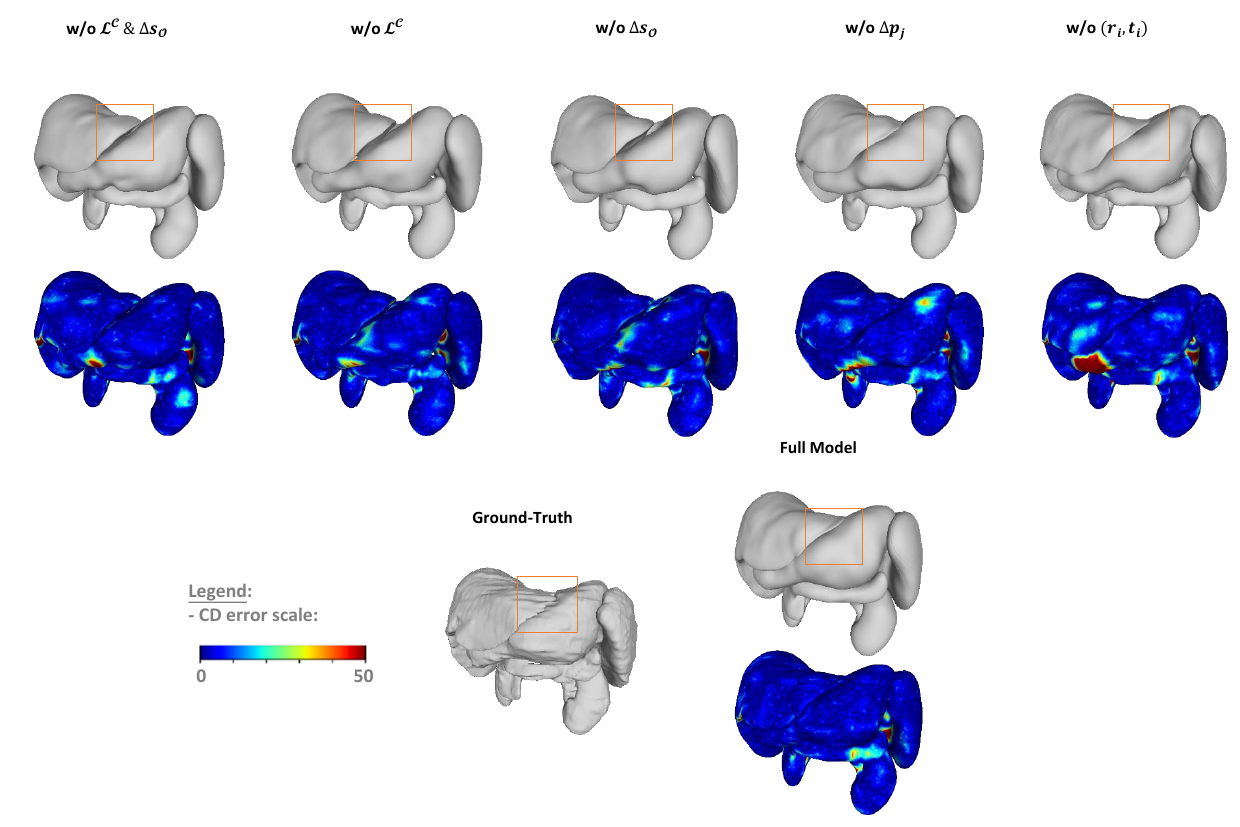}\\
    \caption{Qualitative evaluation of the impact of different methodological components on shape accuracy, in terms of Chamfer Distance (CD), on WORD data \cite{luo2022word}.
    }
    \label{fig:ablation}
\end{figure*}

We qualitatively highlight the impact of different methodological components on shape accuracy in Figure \ref{fig:ablation}. \Eg, we can observe the impact of proposed cross-category mechanism on contact regions between the liver and stomach.

\subsection{Missing-Organ Recovery}



\begin{figure*}
    \centering 
    \includegraphics[width=0.8\linewidth]{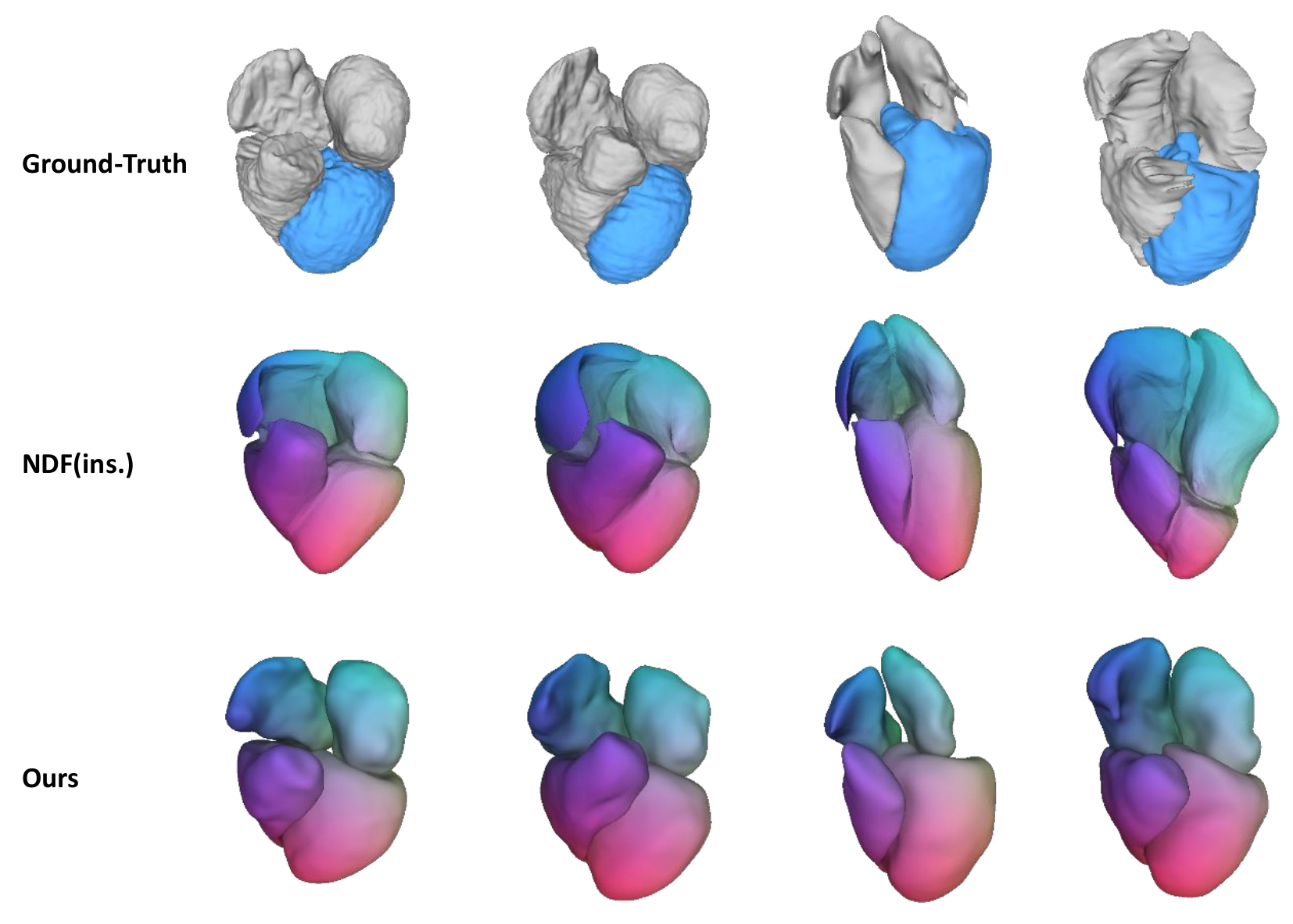}\\
    \caption{Visualization of point correspondences when \texttt{left-myocardium} (blue shape) is missing from instances, on MMWHS data\cite{zhuang2016multi}.
    }
    \label{fig:corres_missing}
\end{figure*}

\paragraph{Point Correspondence for Missing Organ.}
In Figure \ref{fig:corres_missing}, we finally demonstrate how our method can not only recover a plausible and consistent shape for a missing organ, but can also provide dense correspondences for the surface of this extrapolated organ \wrt other instances of this organ.


\subsection{Application to Non-medical Data}
\begin{figure*}
    \centering
    \includegraphics[width = 1.\linewidth]{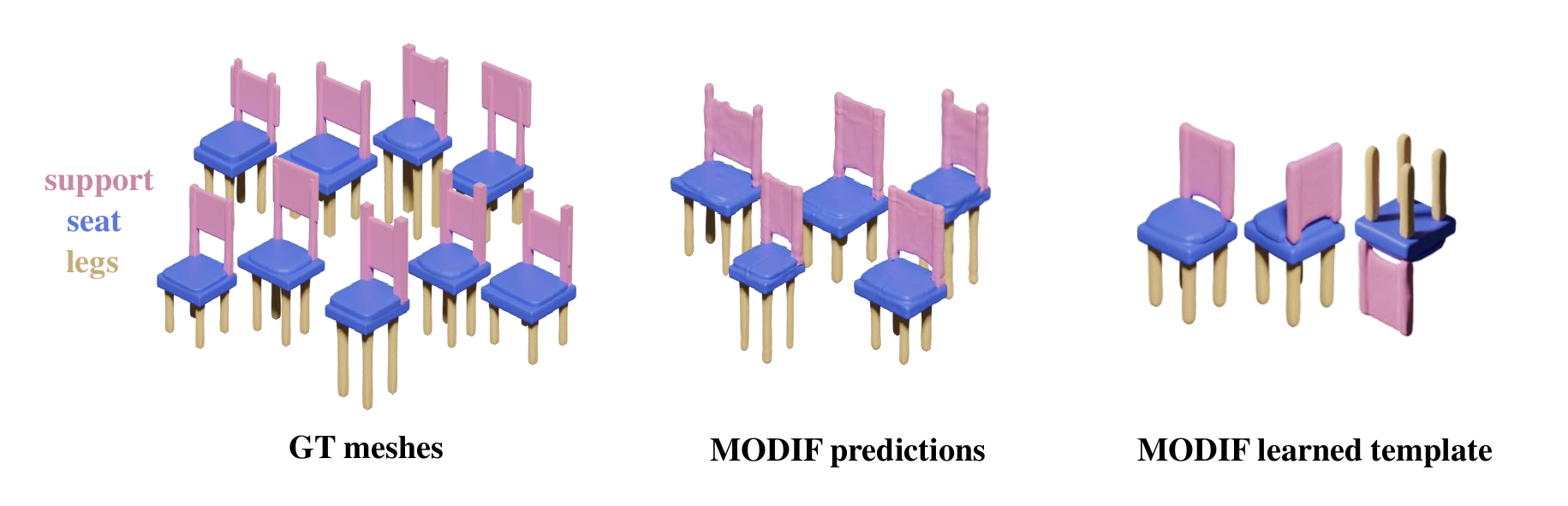}
    \caption{Procedural chairs datasets (3 parts = support, seat, legs) and MODIF results.}
    \label{fig:chairs}
\end{figure*}

While our interest mainly lies in medical applications, we did search the literature for non-medical multi-object benchmarks with topological consistency but could not find any. 
Therefore, we create our own toy dataset of 500 procedurally-generated chairs---composed of three semantic parts (\texttt{seat}, \texttt{legs}, \texttt{back})---and test our method on it. As shown in Figure \ref{fig:chairs}, our method demonstrates the capability to model non-medical multi-part instances.

     

      
    
    

\end{document}